%% file: main.tex
\documentclass[11pt]{article}
\usepackage{arxiv}

\usepackage{xcolor}         %
\usepackage{graphicx}
\usepackage{caption}
\usepackage{subfigure}
\usepackage{tikz}
\usepackage{pgf}
\usepackage{wrapfig}
\usepackage{hyperref}
\usepackage{cleveref}          %
\usepackage{natbib}
\usepackage{color-edits}
\usepackage{booktabs}

\begin{document}

\title{CloudTracks: A Dataset for Localizing \\Ship Tracks in Satellite Images of Clouds}

\author{
    Muhammad Ahmed Chaudhry\thanks{Equal contribution.}\renewcommand{\thefootnote}{\arabic{footnote}}\footnotemark[1] \\
    \texttt{mahmedch@stanford.edu}
    \And
    Lyna Kim\footnotemark[1]\renewcommand{\thefootnote}{\arabic{footnote}}\footnotemark[1] \\ 
    \texttt{lynakim@stanford.edu}
    \And
    Jeremy Irvin\footnotemark[1]\renewcommand{\thefootnote}{\arabic{footnote}}\footnotemark[1] \\ 
    \texttt{jirvin16@cs.stanford.edu}
    \And
    Yuzu Ido\renewcommand{\thefootnote}{\arabic{footnote}}\footnotemark[1] \\ 
    \texttt{yuzu@stanford.edu}
    \And
    Sonia Chu\renewcommand{\thefootnote}{\arabic{footnote}}\footnotemark[1] \\ 
    \texttt{chush@stanford.edu} 
    \And
    Jared Thomas Isobe\renewcommand{\thefootnote}{\arabic{footnote}}\footnotemark[1] \\ 
    \texttt{jtisobe@stanford.edu}
    \AND
    Andrew Y. Ng\renewcommand{\thefootnote}{\arabic{footnote}}\footnotemark[1] \\ 
    \texttt{ang@cs.stanford.edu}
    \And
    Duncan Watson-Parris\renewcommand{\thefootnote}{\arabic{footnote}}\footnotemark[2] \\ 
    \texttt{dwatsonparris@ucsd.edu}
}

\maketitle

\footnotetext[1]{Stanford University}
\footnotetext[2]{UC San Diego, Scripps Institution of Oceanography and Halıcıoğlu Data Science Institute}

\begin{abstract}%
\input{00_abstract}
\end{abstract}


\input{01_intro}

\input{02_data}
\input{03_experiments}
\input{04_results}
\input{05_discussion}
\input{06_conclusion}
\newpage
\input{07_appendix}

\newpage

\vskip 0.2in
\bibliographystyle{plainnat}
\bibliography{refs}

\end{document}

%% file: 00_abstract.tex
Clouds play a significant role in global temperature regulation through their effect on planetary albedo. Anthropogenic emissions of aerosols can alter the albedo of clouds, but the extent of this effect, and its consequent impact on temperature change, remains uncertain. Human-induced clouds caused by ship aerosol emissions, commonly referred to as ship tracks, provide visible manifestations of this effect distinct from adjacent cloud regions and therefore serve as a useful sandbox to study human-induced clouds. However, the lack of large-scale ship track data makes it difficult to deduce their general effects on cloud formation. Towards developing automated approaches to localize ship tracks at scale, we present CloudTracks, a dataset containing 3,560 satellite images labeled with more than 12,000 ship track instance annotations. We train semantic segmentation and instance segmentation model baselines on our dataset and find that our best model substantially outperforms previous state-of-the-art for ship track localization (61.29 vs. 48.65 IoU). We also find that the best instance segmentation model is able to identify the number of ship tracks in each image more accurately than the previous state-of-the-art (1.64 vs. 4.99 MAE). However, we identify cases where the best model struggles to accurately localize and count ship tracks, so we believe CloudTracks will stimulate novel machine learning approaches to better detect elongated and overlapping features in satellite images. We release our dataset openly at {zenodo.org/records/10042922}.

%% file: 01_intro.tex
\section{Introduction}\label{sec:intro}

While anthropogenic greenhouse gasses are primarily responsible for the warming we have experienced to date \citep{IPCC}, some of that warming has been masked by the cooling effect of aerosol \citep{bellouin2020bounding}. These microscopic particles, such as soot and sulfate, directly reflect some solar radiation back to space \citep{10.2307/519399.Ångström} but can also affect the albedo of clouds, making them reflect more sunlight to space \citep{twomey1968comments}. As humans mitigate these emissions, as is necessary to improve air-quality \citep{cohen2017estimates,shindell2019climate}, this cooling effect will be removed, leading to increased warming. The magnitude of this masking effect, and hence the additional warming, is one of the leading uncertainties in future climate change \citep{10.1038/s41558-022-01516-0}.

This uncertainty is due, in part, to the difficulty of observing isolated effects of anthropogenic aerosols on clouds, as they can be confounded with natural meteorological phenomena or other aerosols that are not easily mapped to a specific human activity. Ship emissions release sulfate aerosols that act as cloud condensation nuclei and can enhance cloud droplet numbers, generating \textit{ship tracks}—long, thin trails of enhanced cloud brightness that were observed in some of the very first satellite images of Earth~\citep{10.1175/1520-0469(1966)}. Given their unambiguous source and relatively clean background environment, these features provide a unique opportunity to study human-cloud interaction away from confounding aerosol sources \citep{christensen2022opportunistic}. Previous works have used ship tracks as a sandbox to advance knowledge of anthropogenic aerosol effects \citep{10.1126/science.245.4923.1227, 10.1038/nature03174,10.1029/2019GL084700,10.1029/2010JD014638,10.1175/2009JAS2951.1,10.1029/2012JD017981}.

\begin{figure}
  \centering
  \subfigure[Single Instance]{
    \centering
    \includegraphics[width=0.22\textwidth, height=5cm]{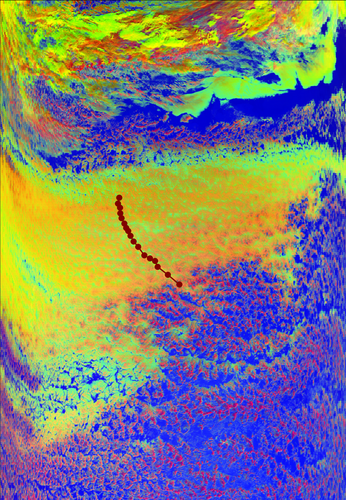}
  }
  \subfigure[Intersecting Instances]{
    \centering
    \includegraphics[width=0.22\textwidth, height=5cm]{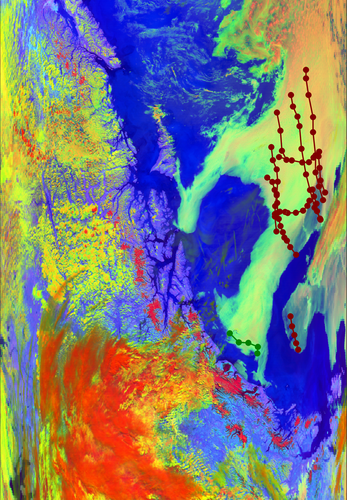}
  }
  \subfigure[Dense Instances]{
    \centering
    \includegraphics[width=0.22\textwidth, height=5cm]{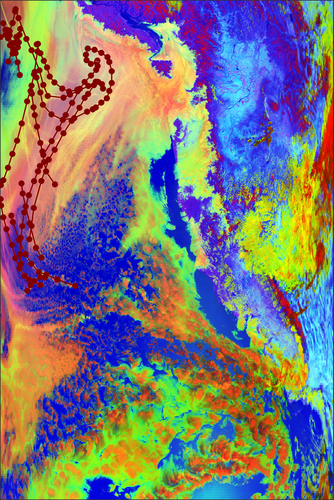}
  }
  \subfigure[No Instances]{
    \centering
    \includegraphics[width=0.22\textwidth, height=5cm]{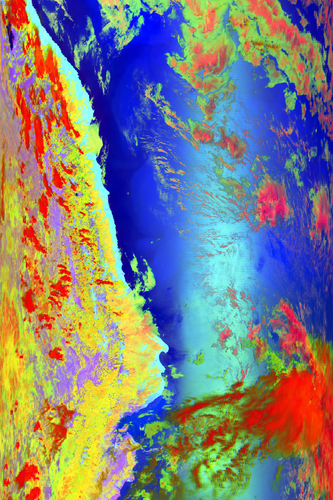}
  }
  \caption{Example images in CloudTracks with ship track instance annotations overlaid. Images in the dataset can contain a single ship track instance (a), intersecting ship track instances (b), densely packed ship track instances (c), and no instances (d).}
  \label{fig:hard_shiptracks}
\end{figure}

Recent works have developed approaches for automatically identifying ship tracks in satellite imagery, in particular using deep learning to detect such cloud formations in satellite images of stratocumulus clouds \citep{watson2022shipping,10.1126/sciadv.abn7988}. Although these preliminary models show promise, their ability to precisely detect ship tracks is still limited. This is primarily due to the physical characteristics of ship tracks as seen in Figure~\ref{fig:hard_shiptracks} that affect their discernibility in satellite imagery: they are long and thin, they are often overlapping and densely packed, they often have sharp turns and kinks in their paths, and they can easily be confused with natural cloud boundaries. Furthermore, previous approaches cannot differentiate specific instances of ship tracks, which limits the ability to accurately assess the impact of tracks on cloud morphology. Existing training data for developing these approaches is limited, which hinders the research community from addressing these challenges.

To this end, we present CloudTracks, a dataset for automatically localizing ship tracks in satellite images of clouds. CloudTracks contains $1,780$ images collected by the Moderate Resolution Imaging Spectroradiometer (MODIS) each hand-labeled with individual ship track instances. This dataset offers significant value compared to previous data \citep{watson2022shipping} for the following reasons. First, CloudTracks is labeled with individual instance annotations rather than binary segmentation masks, which enable the development of instance segmentation models, as we show in this work. These models can be used to identify individual ship tracks at scale, which can allow for isolating the effect of individual ship tracks on cloud formation and enable downstream tasks such as tracking ship movements over time. Second, CloudTracks contains more accurate labels by employing a systematic annotation procedure, as detailed in \Cref{sec:data}. We show that this leads to improved modeling performance both qualitatively and quantitatively. Third, CloudTracks contains additional images with no ship tracks to help reduce the false positive detections found in prior work \citep{watson2022shipping}.

We develop semantic segmentation and instance segmentation models on the dataset and show that the models achieve a new state-of-the-art in ship track localization. However, we find that there are challenging aspects of the ship track detection task that hinder the ability of well-established segmentation models. First, ship tracks often manifest as long and thin objects with sharp turns and kinks, which can be difficult to localize contiguously. Second, the tracks often overlap, sometimes in dense clusters, which leads to challenges with differentiating individual instances. We hope that CloudTracks helps facilitate the development of methods to solve these challenges and helps enable more accurate ship track detection models that can further our understanding of anthropogenic effects on clouds and the climate.

%% file: 02_data.tex
\section{Data}\label{sec:data}
\begin{table}[t]
  \centering
  \begin{tabular}{cccc}
    \toprule
    \textbf{MODIS Channel} & \textbf{Wavelength} & \textbf{Spectrum} & \textbf{Physical Characteristic} \\
    \midrule
    1 & 645nm & Visible & Optical Thickness \\
    20 & 3.75 micrometers & Near Infrared & Cloud Droplet Size \\
    32 & 12.5 micrometers & Infrared & Cloud Temperature \\
    \bottomrule
  \end{tabular}
  \vskip 0.1in
  \caption{Characteristics of the channels used in the MODIS false color composite satellite imagery.}
  \label{tab:channels}
\end{table}

\subsection{MODIS Images}
CloudTracks contains 1,780 NASA MODIS Terra and Aqua images
collected between 2002 and 2021 inclusive over the various stratocumulus cloud regions (such as the East Pacific and East Atlantic) where ship tracks have commonly been observed. Each image has a dimension 1354 x 2030 and a spatial resolution of 1km. For processing images, we followed the methodology described in \citep{watson2022shipping}. Of the 36 bands collected by the instruments, we selected channels 1, 20, and 32 to capture useful physical properties of cloud formations, as listed in Table~\ref{tab:channels}. We applied histogram equalization to scale the channels and construct false color composite images to be used in the final dataset. 

\subsection{Labeling Methodology}\label{sec:data.labeling}
We manually labeled each image in the dataset using a widely used open-source polygonal annotation tool \citep{Wada_Labelme_Image_Polygonal}. We used the following criteria to confirm whether an object was a ship track: the object had to be quasi-linear that gradually evolved into a more diffuse track, and the track had to be continuous above background cloud formation. Each ship track instance was annotated as a sequence of points starting from the head of the track (if visible) and ending at the last discernible portion of the track. To handle cases where the end of the track was unclear, we extended the track as far as possible until it was clear the ship track had terminated. We continued the track through occlusions if there was a clear track before and after the occlusion with similar directionality and shape. We began with annotations from a previous work \citep{watson2022shipping}, removed or modified any annotations that did not conform to the above labeling criteria and procedure, and annotated any new ship tracks not already labeled in the original set of annotations.
We employed the following strategies to increase the quality of the ship track labels. To increase inter-rater agreement, a calibration set of 125 images was used to create clear definitions of ship tracks and reduce annotation differences on challenging cases. To further capture the possible subjectivity of determining ship tracks, we created two classes for labels, “ship track” and “uncertain.”  We labeled objects entirely consisting of borderline diffuseness as uncertain. Although investigating the impact of including the “uncertain” ship tracks is an interesting research direction, for all subsequent experiments we only retain the “ship track” labels. Examples of images with ship track annotations overlaid are shown in Figure~\ref{fig:hard_shiptracks}.

\subsection{Dataset Preprocessing}
We converted the ship track instance annotations to semantic segmentation and instance segmentation masks using the following procedures. For semantic segmentation, adjacent points within each instance were joined with a line segment and then buffered using a pixel width of 10. The buffered line segments from all instances were then rasterized to create a binary segmentation mask with the same dimensions as the corresponding MODIS image. For instance segmentation, each instance was converted to a polygon with a pixel width of 10 for consistency with semantic segmentation and each instance polygon was converted to pixel coordinates relative to the dimensions of the corresponding MODIS image.
Before inputting the images into the models, we cropped the original 1354 x 2030 images to obtain two images with 1354 x 1015 resolution. The binary masks obtained from the rasterization and the polygonized instance annotations were cropped in the same way to obtain the corresponding ground truth labels. This cropping was selected to balance the benefits of increased context with GPU memory constraints.
Moreover, our preliminary experiments showed that using crops smaller than 1354 x 1015 indeed harmed model performance since a smaller footprint prohibits the models from capturing sufficient context in the images to accurately localize the ship tracks. 
The resulting dataset consists of 3,560 image and mask pairs.  
We randomly split the dataset into a training set (70\%), validation set (20\%), and test set (10\%). Statistics of the images and annotations in each split are shown in Table~\ref{tab:train_val_test_split}.

\begin{table}[t!]
  \centering
  \begin{tabular}{lcccc}
    \toprule
        & Training & Validation     & Test & Total\\
    \midrule
    Positive Images & 1,433 & 415 & 205 & 2,053  \\
    Negative Images & 1,066 & 305 & 136 & 1,507  \\
    Total Images  & 2,499 & 720  & 341 & 3,560      \\
    Ship Track Instances & 8,786 & 2,568 & 1,141 & 12,495\\
    \bottomrule
  \end{tabular}
  \vskip 0.1in
  \caption{Statistics of the training, validation, and test sets in CloudTracks.}
  \label{tab:train_val_test_split}
\end{table}

%% file: 03_experiments.tex
\section{Experiments}\label{sec:experiments}
We ran several experiments on CloudTracks assessing the ship track localization and counting performance of semantic segmentation and instance segmentation models. For all models, we tuned the learning rate (1e-3, 1e-4, 1e-5) and optimizer (Adam with default parameters, SGD with a momentum of 0.9) with weight decay of 0.0001 and a batch size of 2. We ran each experiment three times with different random seeds and report the mean and standard deviation of the results. All experiments were run on a single NVIDIA A4000 GPU.

\subsection{Semantic Segmentation}
We developed a variety of semantic segmentation models which input a satellite image and produce a per-pixel classification indicating which pixels in the image correspond to ship tracks. Before inputting the images into the models, we resized the images to a 672 x 672 resolution to adhere to segmentation architecture requirements and memory constraints. We used spatial augmentations including random horizontal and vertical flips (each with 50\% chance), affine scaling along the x and y axes (up to 95\% to 105\% of the original image sizes), affine translations (by -30\% to +30\% on the x and y axes independently), and rotations of 90, 180, or 270 degrees. Furthermore, we experimented with different loss functions including jaccard loss, binary cross entropy (BCE), and a convex combination of the two with equal weight.
Our preliminary experiments explored well-established semantic segmentation model architectures including DeepLabV3 \citep{chen2017rethinking}, UNet \citep{ronneberger2015u}, and Feature Pyramid Networks (FPNs) \citep{lin2017feature} as well as various backbone architectures pre-trained on ImageNet \citep{deng2009imagenet} including ResNets (ResNet18, ResNet34, ResNet50, ResNet101, ResNet152) \citep{he2016deep}, ResNeXt (ResNeXt101)  \citep{xie2017aggregated}, DenseNets (DenseNet121 and DenseNet161) \citep{huang2017densely}, and EfficientNet (EfficientNet-b7) \citep{tan2019efficientnet}.  The best model was a UNet architecture with a EfficientNet-b7 backbone trained with a learning rate of 1e-4 with an Adam optimizer. We refer to this as the ``Best Semantic Segmentation'' model. We compared this to a reimplementation of the model in \cite{watson2022shipping} which uses a UNet architecture with a ResNet152 backbone trained using an Adam optimizer with a learning rate of 1e-2 on the original, uncorrected labels of the dataset using the same splits. We evaluated both models on the test set with the corrected labels.

\subsection{Instance Segmentation}
We developed instance segmentation models which input a satellite image and generate a per-pixel classification indicating which pixels in the image correspond to ship tracks as well as bounding boxes identifying separate instances of ship tracks. During training, we used common data augmentations for instance segmentation including random resizing to different image scales (1333 x 800, 1333 x 768, 1333 x 736, 1333 x 704, 1333 x 672, and 1333 x 640), random horizontal and vertical flips (each with a 50\% flip ratio), and padding (with a size divisor of 32).
We experimented with two instance segmentation architectures, namely Mask-RCNN \citep{he2017mask} and SOLOv2 \citep{wang2020solov2} which uses an FPN with deformable convolutions \citep{zhu2019deformable} and we explored the use of two ResNet backbone encoders (ResNet50, ResNet101) pre-trained on ImageNet. We used the default loss function for each model architecture, namely cross entropy loss for the class and mask losses and L1 for the bounding box regression in Mask-RCNN, and dice loss for the mask loss and focal loss for the class loss with SOLOv2. The best model was a SOLOv2 architecture with a ResNet101 backbone trained with a learning rate of 1e-3 with an SGD optimizer, and we refer to this model as the ``Best Instance Segmentation'' model.

\subsection{Evaluation}
\subsubsection{Localization}
For both tasks, we evaluated the localization performance of the models using Intersection over Union (IoU) on the images with ship tracks and pixel-level precision, recall, F1 score, and specificity on all images. However, we observed that ship track predictions can be high quality but still attain low values of these metrics as ship tracks can be very narrow (Figure~\ref{fig:buffering}) and the ground truth annotations are not always perfectly precise.
To address this, we evaluated with more relaxed performance metrics in the following way. We considered any predicted ship track pixels within $N$ pixels of the annotated ship tracks as true positives rather than false positives and any missed annotated ship track pixels within $N$ pixels of the predicted ship tracks as true positives rather than false negatives. Then we computed all metrics in the usual way. We set $N=5$ because we use a width of 10 pixels when generating the ship track annotations. We report the relaxed metrics in all subsequent experiments, and report the original metrics for comparison in Table~\ref{tab:metric-table-unbuffered} of the Appendix. 

\subsubsection{Instance Counting}
We measured the effectiveness of the models to count the number of instances using mean average error (MAE) between the predicted number and true number of ship tracks in each image. To obtain instance predictions with the semantic segmentation models, we followed \cite{watson2022shipping} and used contouring to detect ship track polygons in the predicted mask. For the baseline semantic segmentation model, we set the confidence level of the contouring using their setting of 0.8. For our semantic segmentation model, we tuned the confidence level of the contouring based on MAE on the validation set and found that a 0.6 level works best.  We also tuned the confidence threshold for determining whether to keep or drop bounding boxes produced by the instance segmentation by using the MAE on the validation set, and found that using a threshold of 0.2 achieves the highest MAE (as well as IoU). These settings were used for the models respectively when evaluating on the test set.

%% file: 04_results.tex
\section{Results}\label{sec:results}

\begin{table}[!t]
  \centering
  \resizebox{\columnwidth}{!}{%
  \begin{tabular}{l|ccccc}
    \toprule
    Model     & IoU & Precision & Recall & F1 & Specificity \\
    \midrule
    \cite{watson2022shipping} & 48.65 $\pm$ 2.28 & 63.57 $\pm$ 3.06 & 49.55 $\pm$ 3.74 & 27.83 $\pm$ 1.66 & 99.88 $\pm$ 0.01 \\
    Best Semantic Segmentation  & 61.29 $\pm$ 1.11 & 77.63 $\pm$ 1.29  & 71.69 $\pm$ 1.98  & 37.26 $\pm$ 0.27 & 99.89 $\pm$ 0.01    \\
    Best Instance Segmentation  & 57.55 $\pm$ 0.24 & 74.26 $\pm$ 1.02  & 67.84 $\pm$ 0.71  & 35.45 $\pm$ 0.06 & 99.88 $\pm$ 0.01  \\ 
    \bottomrule
  \end{tabular}%
}
  \vskip 0.1in
  \caption{Test set performance metrics of the semantic and instance segmentation models trained on CloudTracks. Error bars represent the standard deviation of three identical runs with different random seeds.}
  \label{tab:metric-table}
\end{table}

    \begin{figure}[ht!]
    \setlength{\tabcolsep}{4pt}
        \centering
        \begin{tabular}{ccccc}
         \includegraphics[width=80pt, height=80pt]{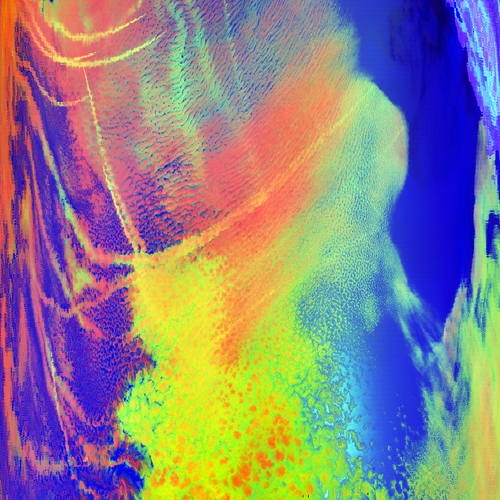}    & \includegraphics[width=80pt, height=80pt]{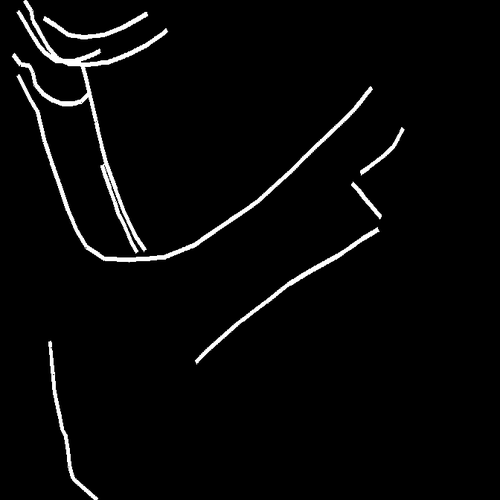} &
        \includegraphics[width=80pt, height=80pt]{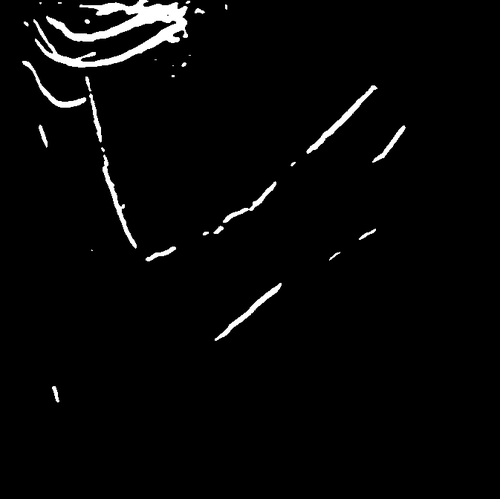} &
        \includegraphics[width=80pt, height=80pt]{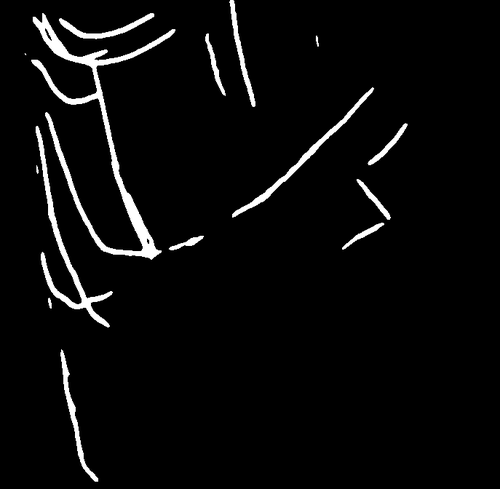} &
        \includegraphics[width=80pt, height=80pt]{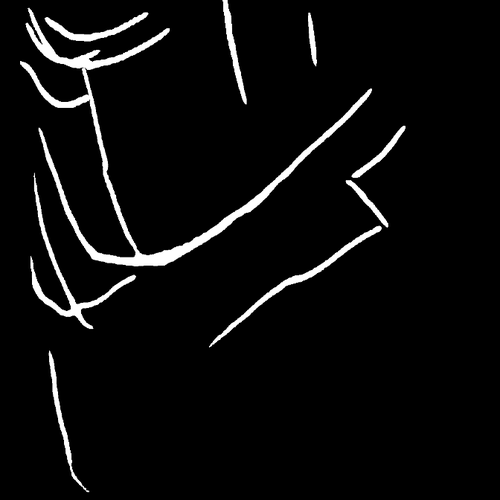}\\
        \includegraphics[width=80pt, height=80pt]{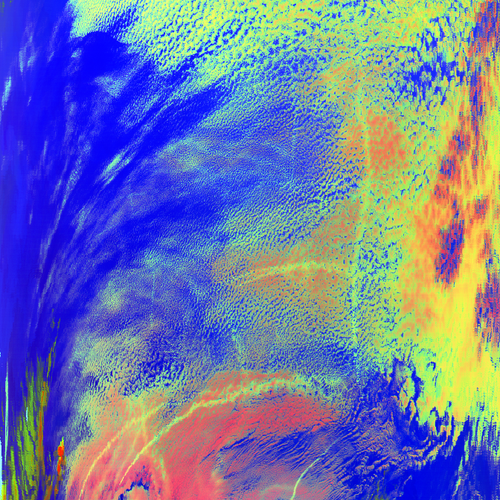} &
        \includegraphics[width=80pt, height=80pt]{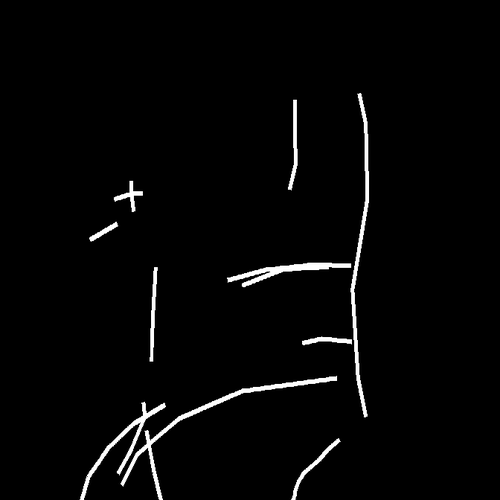} &
        \includegraphics[width=80pt, height=80pt]{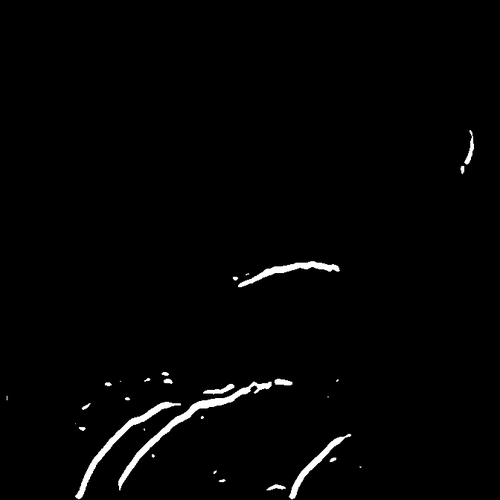} &
        \includegraphics[width=80pt, height=80pt]{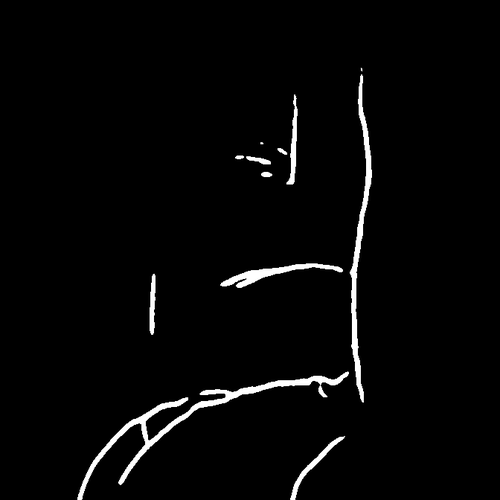} &
        \includegraphics[width=80pt, height=80pt]{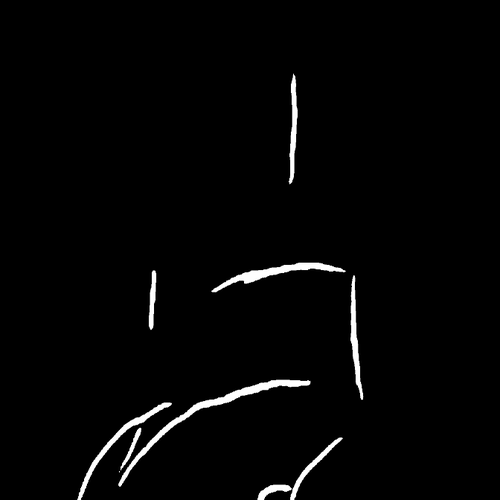}\\
        \includegraphics[width=80pt, height=80pt]{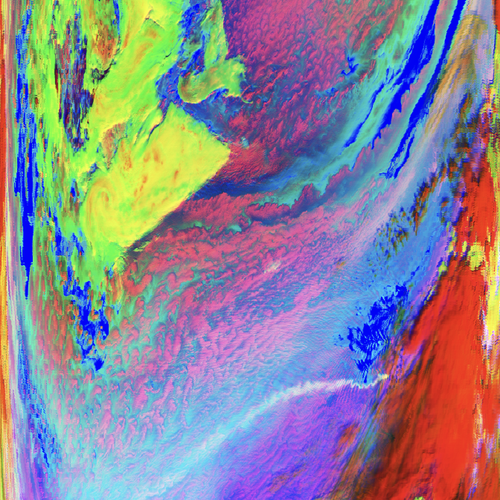} &
        \includegraphics[width=80pt, height=80pt]{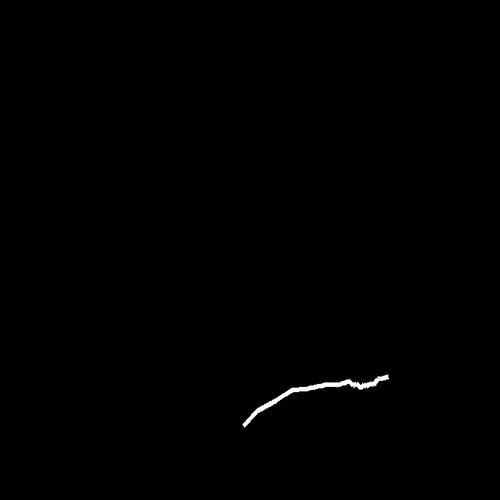} &
        \includegraphics[width=80pt, height=80pt]{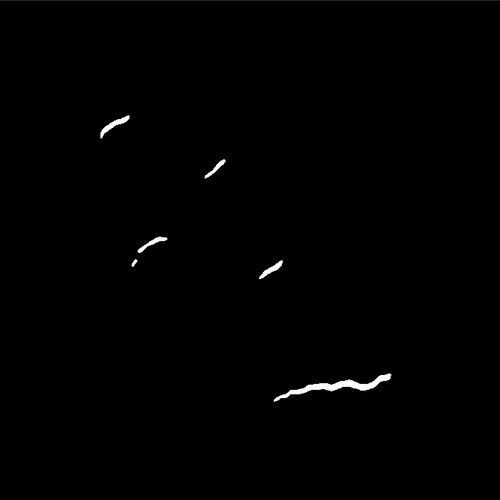} &
        \includegraphics[width=80pt, height=80pt]{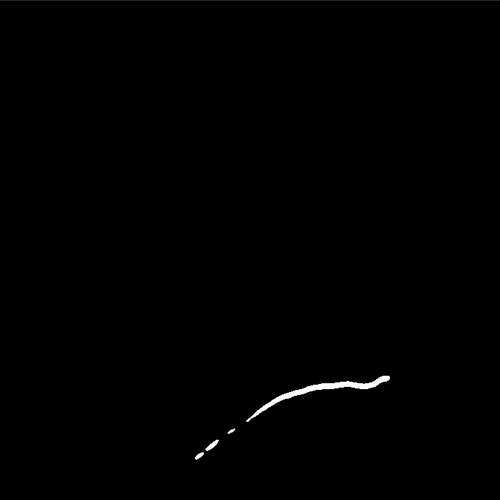} &
        \includegraphics[width=80pt, height=80pt]{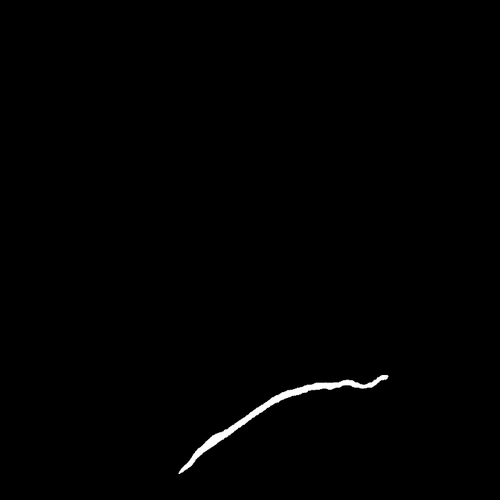}
        \end{tabular}
        \caption{Example images in the test set showcasing the patterns of improvements of our models over the baseline semantic segmentation model. The images from left to right are: False-colored satellite image, Ground-truth annotation, \cite{watson2022shipping} prediction, Best Semantic Segmentation prediction, and Best Instance Segmentation prediction. The first row shows more contiguous tracks predicted by the improved models, second row shows an example of increased sensitivity to ship tracks, and third shows less extraneous predictions.} 
        \label{fig:semseg_qual}
    \end{figure}

\subsection{Localization}
Our semantic and instance segmentation models performed comparably on ship track localization and substantially outperformed the baseline model in \citep{watson2022shipping} across all measured evaluation metrics (Table~\ref{tab:metric-table}). Specifically, the best semantic segmentation model outperformed the baseline by 12.64 IoU and 9.43 F1. Our best instance segmentation model underperformed the best semantic segmentation model across all metrics, notably by 3.74  IoU and 1.81 F1. This may be because the semantic segmentation model was solely optimized for localization whereas the instance segmentation model was jointly optimized for localization and instance detection. Still, the localization performance of the instance segmentation model was close to that of the best segmentation model. 
Predictions of both models were strong qualitatively compared to the baseline (Figure~\ref{fig:semseg_qual}). Both models produced longer, more continuous tracks and fewer spurious short predictions (see the first row of the figure). Both models were more sensitive to ship tracks, detecting many more of the tracks in the images (for example the vertical ship tracks in the second row). The increased sensitivity did not come at the cost of lower precision; both of our models made fewer extraneous predictions than the baseline which often made false positive predictions due to confounding cloud features (see third row).

\subsection{Instance Counting}
\begin{figure}[ht!]
     \centering
     \subfigure[\cite{watson2022shipping}]{
         \includegraphics[width=0.3\textwidth]{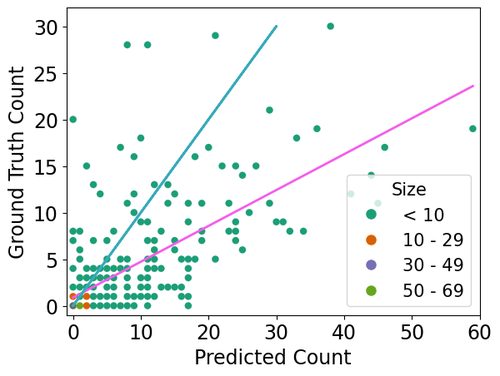}
         \label{fig:baseline_counts}
     }
     \subfigure[Best Semantic Segmentation]{
         \includegraphics[width=0.3\textwidth]{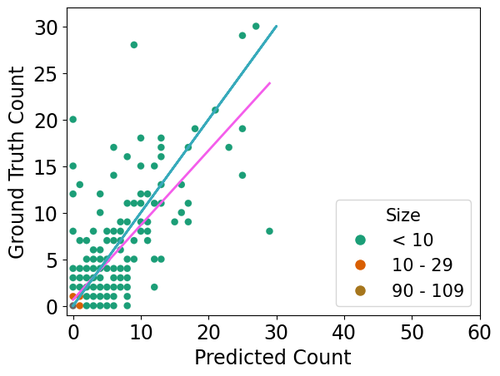}
         \label{fig:semseg_counts}
      }
     \subfigure[Best Instance Segmentation]{
         \includegraphics[width=0.3\textwidth]{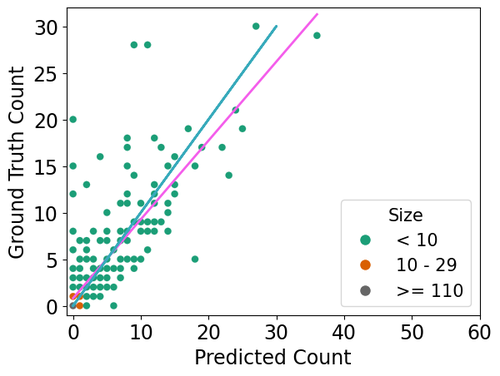}
         \label{fig:iseg_counts}
      }
        \caption{Instance counting performance of the three segmentation models on the CloudTracks test set. The optimal line is shown in teal and best fit line between the model predicted counts and ground truth counts is shown in pink.}
        \label{fig:s_counts}
\end{figure}

\begin{figure}[!ht]
     \centering
     \subfigure[Long track detections]{         \includegraphics[width=0.47\textwidth]{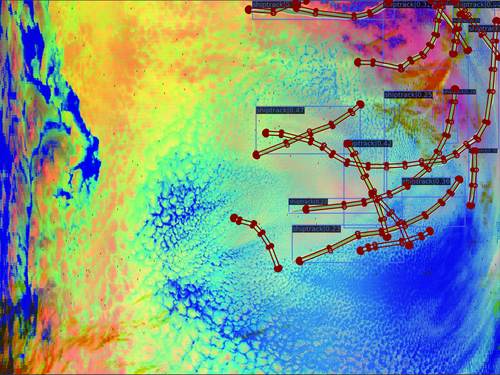}
         \label{fig:iseg_local_1}}
     \subfigure[Dense tracks]{         \includegraphics[width=0.47\textwidth]{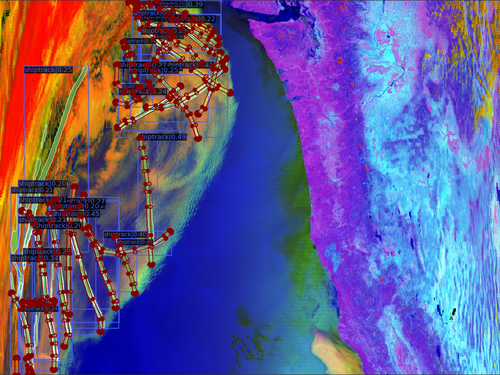}
         \label{fig:iseg_local_5}
     }
     \vskip\baselineskip
     \subfigure[Difficulty of crossings]{ \includegraphics[width=0.47\textwidth]{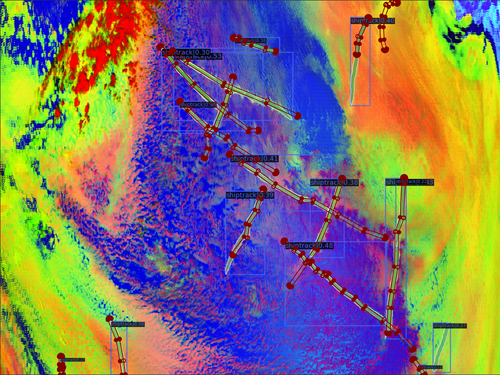}
            \label{fig:iseg_local_3}
    }
    \subfigure[Ship track mimicking cloud features]{ \includegraphics[width=0.47\textwidth]{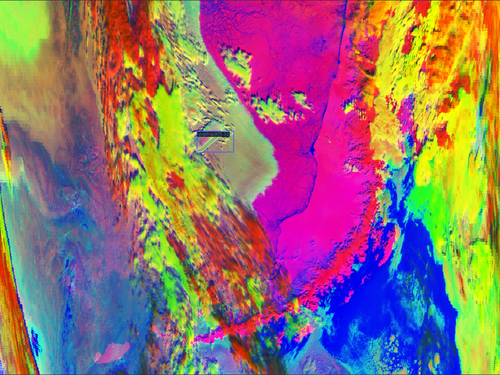}  
            \label{fig:iseg_qual_4}
    }

    \caption{Example instance segmentation predictions on the test set. Ground truth ship track polygons are shown in red and ship track predicted masks are shown in green with corresponding predicted bounding box and confidence score in blue. The model often successfully identified and localized instances of long ship tracks (a) and densely packed tracks (b). Common mistakes included difficulties identifying single instances in cases where many ship tracks cross and images with features that mimic ship tracks, leading to false positive predictions.}
    \label{fig:iseg_quals}
\end{figure}

The instance segmentation model performed best in identifying the correct number of instances (Figure~\ref{fig:s_counts}).  Specifically, the instance segmentation model achieved an MAE of 1.64 $\pm$ 0.05 compared to the best semantic segmentation model's performance of 1.90 $\pm$ 0.11 MAE and baseline of 4.99 $\pm$ 0.54 MAE. The baseline semantic segmentation model tended to predict too many ship tracks compared to the ground truth, in some cases falsely identifying more than 40 ship tracks. It also often produced a nonzero amount of predictions on images without any ship tracks. The best semantic segmentation and instance segmentation models, however, tended to slightly underpredict the amount of ship tracks in the image. Both models did not ever predict more than 40 ship tracks in an image, and the instance segmentation model almost always identified the number of ship tracks within 10 tracks of the ground truth. Both models also produced less false positive predictions on images without ship tracks, with the instance segmentation model slightly outperforming the semantic segmentation model on those images. 
Representative examples of correct and incorrect instance predictions are shown in Figure~\ref{fig:iseg_quals}. The model often accurately localized long ship tracks and was able to identify individual ship tracks in densely packed groups. The model struggled with images that contain lots of overlap and crossings between ship tracks. The model also produced some false positive predictions on features of clouds that appear similarly to ship tracks.

%% file: 05_discussion.tex
\section{Discussion}\label{sec:discussion}
By leveraging CloudTracks, we achieved state-of-the-art ship track semantic and instance segmentation performance. In addition to a ship track localization improvement of 8.90 IoU and 18.29 F1 compared to previous work, we improved upon ship track detection for key qualitative metrics, such as prediction continuity and precision in difficult satellite images with confusing objects. The additional granularity of separate ship track instances makes the resulting dataset and model outputs substantially more useful for climate science research. 

\subsection{Technical Challenges}
Although the models we train on CloudTracks demonstrated superior performance to prior methods, the dataset still presents multiple challenges to the segmentation models. Thin, long, and continuous objects like ship tracks are uncommon in large-scale image datasets used for pretraining, such as COCO \citep{lin2014microsoft}. This may explain why model architectures developed for that dataset, and subsequent pretrained models fine-tuned on CloudTracks, struggle to achieve excellent performance on CloudTracks. We found that this manifests as discontinuous predictions and difficult localizing instances at crossings and occlusions where ship tracks are densely packed. It is worthwhile for future work to explore improving model performance using methods developed to address discontinuity issues including approaches from boundary detection \citep{wang2022active} and deformable linear object detection \citep{keipour2022detection}, as well as methods to handle dense and overlapping objects \citep{goldman2019precise,chen2022instance}. Such approaches could also help improve related climate tasks such as detecting contrails~\citep{ng2023opencontrails}.

\subsection{Dataset Limitations}
\paragraph{Geographic Diversity}
As CloudTracks only contains satellite images in the Pacific Ocean, models trained on CloudTracks may not generalize to new geographic locations not represented in the dataset. Future work may benefit from obtaining satellite images across a more globally distributed area. 

\paragraph{Label Consistency}
It can be difficult to consistently identify ship tracks in satellite imagery, largely due to subjective judgements about when ship tracks terminate. While we tried to maximize inter-rater agreement using clear rules for annotation and an additional ``uncertain'' class to capture particularly difficult cases, we acknowledge that the labels in CloudTracks still may not be perfect. We suggest that users of the dataset are cautious about this, and although we did not explore it in this work, we believe the use of the uncertainty labels are also an interesting direction for future work.
Furthermore, the ship track labels were generated as fixed-width buffers of the sequences of points, but we observed that ship tracks often manifest as varying-width objects, commonly starting from a narrow point then widening outward through diffusion later in the track. We used relaxed evaluation metrics to address this, but users of the dataset may want to explore the use of different ship track widths which is straightfoward to do from the released version of CloudTracks.

\paragraph{Climate Impacts}
While our models and experiments are lightweight, we acknowledge the carbon footprint associated with training deep learning models \citep{schwartz2020green}. We recommend that future researchers are judicious about energy usage when training their own models on CloudTracks and in general \citep{heguerte2023estimate}. 

%% file: 06_conclusion.tex
\section{Conclusion}\label{sec:conclusion}
We introduce CloudTracks, a new dataset containing 3,560 satellite images hand-annotated with more than 12,000 ship track instances. We benchmarked semantic and instance segmentation experiments on the dataset and find that they achieve state-of-the-art performance on ship track localization. We hope the dataset will stimulate novel machine learning approaches to improve detection of thin, occluded, and intersecting objects in noisy geospatial imagery, as well as advance research about anthropogenic aerosol effects on clouds and climate change. 

%% file: 07_appendix.tex
\section*{Appendix A}\label{sec:appendix}

\begin{table}[!ht]
  \centering
  \resizebox{\columnwidth}{!}{%
  \begin{tabular}{l|ccccc}
    \toprule
    Model     & IoU & Precision & Recall & F1 & Specificity \\
    \midrule
    \cite{watson2022shipping} & 17.30 $\pm$ 1.22 & 41.06 $\pm$ 1.40 & 31.30 $\pm$ 2.86 & 17.75 $\pm$ 1.18 & 99.80 $\pm$ 0.01 \\
    Best Semantic Segmentation  & 26.44 $\pm$ 0.17 & 48.61 $\pm$ 1.02  & 51.10 $\pm$ 0.87  & 24.91 $\pm$ 0.11 & 99.76 $\pm$ 0.01    \\
    Best Instance Segmentation  & 26.64 $\pm$ 0.84 & 49.08 $\pm$ 2.38  & 53.72 $\pm$ 1.21  & 25.62 $\pm$ 0.41 & 99.76 $\pm$ 0.03  \\ 
    \bottomrule
  \end{tabular}%
}
  \caption{Original (unrelaxed) test set performance metrics of the semantic and instance segmentation models trained on CloudTracks. Error bars represent the standard deviation of three identical runs with different random seeds.}
  \label{tab:metric-table-unbuffered}
\end{table}

\begin{figure}[!ht]
    \centering
    \tikzset{every picture/.style={line width=0.75pt}} %
    
    \begin{tikzpicture}[x=0.75pt,y=0.75pt,yscale=-1,xscale=1]
    
    \draw    (80,9.97) -- (79.67,210.5) ;
    \draw    (80,9.97) -- (220,10.97) ;
    \draw    (79.67,210.5) -- (219.67,210.5) ;
    \draw    (220,10.97) -- (219.67,210.5) ;
    \draw    (102.67,45.5) .. controls (142.67,15.5) and (136.67,170.5) .. (176.67,140.5) ;
    \draw    (100.67,119.5) .. controls (140.67,89.5) and (160,142) .. (200,112) ;
    \draw    (251,10.47) -- (250.67,211) ;
    \draw    (251,10.47) -- (391,11.47) ;
    \draw    (250.67,211) -- (390.67,211) ;
    \draw    (391,11.47) -- (390.67,211) ;
    \draw    (273.67,46) .. controls (313.67,16) and (307.67,171) .. (347.67,141) ;
    \draw    (271.67,120) .. controls (311.67,90) and (331,142.5) .. (371,112.5) ;
    \draw [color={rgb, 255:red, 208; green, 2; blue, 27 }  ,draw opacity=1 ][line width=1.5]  [dash pattern={on 5.63pt off 4.5pt}]  (270.67,114.5) .. controls (310.67,84.5) and (331.67,137.5) .. (371.67,107.5) ;
    \draw [color={rgb, 255:red, 208; green, 2; blue, 27 }  ,draw opacity=1 ][line width=1.5]  [dash pattern={on 5.63pt off 4.5pt}]  (276.67,39.5) .. controls (316.67,9.5) and (309.67,165.5) .. (349.67,135.5) ;
    \draw    (420,10.47) -- (419.67,211) ;
    \draw    (420,10.47) -- (560,11.47) ;
    \draw    (419.67,211) -- (559.67,211) ;
    \draw    (560,11.47) -- (559.67,211) ;
    \draw [color={rgb, 255:red, 155; green, 155; blue, 155 }  ,draw opacity=0.38 ][line width=6]    (442.67,46) .. controls (482.67,16) and (476.67,171) .. (516.67,141) ;
    \draw [color={rgb, 255:red, 155; green, 155; blue, 155 }  ,draw opacity=0.38 ][line width=6]    (440.67,120) .. controls (480.67,90) and (500,142.5) .. (540,112.5) ;
    \draw [color={rgb, 255:red, 126; green, 211; blue, 33 }  ,draw opacity=1 ][line width=1.5]  [dash pattern={on 5.63pt off 4.5pt}]  (439.67,114.5) .. controls (479.67,84.5) and (500.67,137.5) .. (540.67,107.5) ;
    \draw    (442.67,46) .. controls (482.67,16) and (476.67,171) .. (516.67,141) ;
    \draw    (440.67,120) .. controls (480.67,90) and (500,142.5) .. (540,112.5) ;
    \draw  [draw opacity=0][fill={rgb, 255:red, 155; green, 155; blue, 155 }  ,fill opacity=0.38 ] (443.17,123.23) .. controls (443.16,123.24) and (443.14,123.25) .. (443.13,123.27) .. controls (441.62,124.88) and (439.07,124.95) .. (437.44,123.42) .. controls (435.81,121.89) and (435.71,119.34) .. (437.22,117.73) .. controls (437.55,117.38) and (437.93,117.1) .. (438.34,116.89) -- cycle ;
    \draw  [draw opacity=0][fill={rgb, 255:red, 155; green, 155; blue, 155 }  ,fill opacity=0.38 ] (514.23,137.74) .. controls (514.24,137.72) and (514.25,137.71) .. (514.26,137.7) .. controls (515.79,136.1) and (518.34,136.05) .. (519.96,137.59) .. controls (521.58,139.13) and (521.65,141.68) .. (520.13,143.28) .. controls (519.8,143.63) and (519.41,143.91) .. (519,144.11) -- cycle ;
    \draw [color={rgb, 255:red, 126; green, 211; blue, 33 }  ,draw opacity=1 ][line width=1.5]  [dash pattern={on 5.63pt off 4.5pt}]  (445.27,41.17) .. controls (485.27,11.17) and (478.27,167.17) .. (518.27,137.17) ;
    \draw  [draw opacity=0][fill={rgb, 255:red, 155; green, 155; blue, 155 }  ,fill opacity=0.38 ] (537.59,109.3) .. controls (537.6,109.28) and (537.61,109.27) .. (537.62,109.26) .. controls (539.14,107.65) and (541.69,107.59) .. (543.32,109.12) .. controls (544.94,110.66) and (545.03,113.2) .. (543.52,114.81) .. controls (543.18,115.17) and (542.8,115.44) .. (542.39,115.65) -- cycle ;
    \draw  [draw opacity=0][fill={rgb, 255:red, 155; green, 155; blue, 155 }  ,fill opacity=0.38 ] (445.12,49.26) .. controls (445.11,49.27) and (445.09,49.28) .. (445.08,49.3) .. controls (443.56,50.9) and (441.01,50.95) .. (439.39,49.41) .. controls (437.77,47.87) and (437.69,45.32) .. (439.21,43.72) .. controls (439.54,43.37) and (439.93,43.09) .. (440.34,42.89) -- cycle ;
        
    \draw (100,223) node [anchor=north west][inner sep=0.75pt]  [font=\small] [align=left] {Ground truth mask};
    \draw (267,223) node [anchor=north west][inner sep=0.75pt]  [font=\small] [align=left] {Predictions overlaid};
    \draw (422,222) node [anchor=north west][inner sep=0.75pt]  [font=\small] [align=left] {Buffered ground truth};
    
    \end{tikzpicture}
    \caption{Example demonstrating the motivation to use buffering when computing relaxed evaluation metrics. Importantly, using the original metrics without buffering, an almost perfect prediction (middle figure) translated by a few pixels achieves a very low IoU score.}
    \label{fig:buffering}
\end{figure}
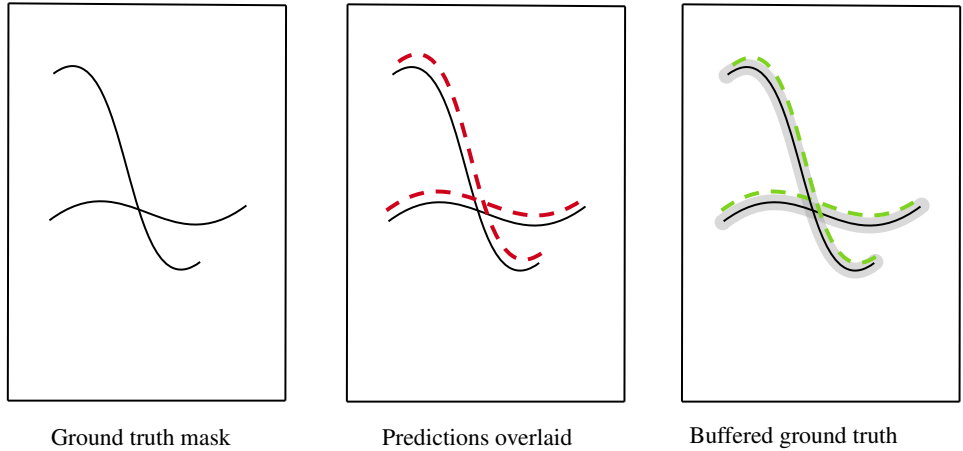

%% file: main.bbl
\begin{thebibliography}{37}
\providecommand{\natexlab}[1]{#1}
\providecommand{\url}[1]{\texttt{#1}}
\expandafter\ifx\csname urlstyle\endcsname\relax
  \providecommand{\doi}[1]{doi: #1}\else
  \providecommand{\doi}{doi: \begingroup \urlstyle{rm}\Url}\fi

\bibitem[Ackerman et~al.(2004)Ackerman, Kirkpatrick, Stevens, and Toon]{10.1038/nature03174}
Andrew~S. Ackerman, Michael~P. Kirkpatrick, David~E. Stevens, and Owen~B. Toon.
\newblock {The impact of humidity above stratiform clouds on indirect aerosol climate forcing}.
\newblock \emph{Nature}, 432\penalty0 (7020):\penalty0 1014--1017, 12 2004.
\newblock ISSN 0028-0836.
\newblock \doi{10.1038/nature03174}.

\bibitem[Albrecht(1989)]{10.1126/science.245.4923.1227}
Bruce~A. Albrecht.
\newblock {Aerosols, Cloud Microphysics, and Fractional Cloudiness}.
\newblock \emph{Science}, 245\penalty0 (4923):\penalty0 1227--1230, 9 1989.
\newblock ISSN 0036-8075.
\newblock \doi{10.1126/science.245.4923.1227}.

\bibitem[Bellouin et~al.(2020)Bellouin, Quaas, Gryspeerdt, Kinne, Stier, Watson-Parris, Boucher, Carslaw, Christensen, Daniau, et~al.]{bellouin2020bounding}
Nicolas Bellouin, Johannes Quaas, Edward Gryspeerdt, Stefan Kinne, Philip Stier, Duncan Watson-Parris, Olivier Boucher, Ken~S Carslaw, Matthew Christensen, A-L Daniau, et~al.
\newblock Bounding global aerosol radiative forcing of climate change.
\newblock \emph{Reviews of Geophysics}, 58\penalty0 (1):\penalty0 e2019RG000660, 2020.

\bibitem[Chen et~al.(2017)Chen, Papandreou, Schroff, and Adam]{chen2017rethinking}
Liang-Chieh Chen, George Papandreou, Florian Schroff, and Hartwig Adam.
\newblock Rethinking atrous convolution for semantic image segmentation.
\newblock \emph{arXiv preprint arXiv:1706.05587}, 2017.

\bibitem[Chen et~al.(2022)Chen, Wu, and Merhof]{chen2022instance}
Long Chen, Yuli Wu, and Dorit Merhof.
\newblock Instance segmentation of dense and overlapping objects via layering.
\newblock \emph{arXiv preprint arXiv:2210.03551}, 2022.

\bibitem[Christensen and Stephens(2011)]{10.1029/2010JD014638}
Matthew~W. Christensen and Graeme~L. Stephens.
\newblock {Microphysical and macrophysical responses of marine stratocumulus polluted by underlying ships: Evidence of cloud deepening}.
\newblock \emph{Journal of Geophysical Research: Atmospheres}, 116\penalty0 (D3), 2 2011.
\newblock ISSN 0148-0227.
\newblock \doi{10.1029/2010JD014638}.

\bibitem[Christensen et~al.(2009)Christensen, Jr., and Tahnk]{10.1175/2009JAS2951.1}
Matthew~W. Christensen, James A.~Coakley Jr., and William~R. Tahnk.
\newblock {Morning-to-Afternoon Evolution of Marine Stratus Polluted by Underlying Ships: Implications for the Relative Lifetimes of Polluted and Unpolluted Clouds}.
\newblock \emph{Journal of the Atmospheric Sciences}, 66\penalty0 (7):\penalty0 2097--2106, 7 2009.
\newblock ISSN 0022-4928.
\newblock \doi{10.1175/2009JAS2951.1}.

\bibitem[Christensen et~al.(2022)Christensen, Gettelman, Cermak, Dagan, Diamond, Douglas, Feingold, Glassmeier, Goren, Grosvenor, et~al.]{christensen2022opportunistic}
Matthew~W Christensen, Andrew Gettelman, Jan Cermak, Guy Dagan, Michael Diamond, Alyson Douglas, Graham Feingold, Franziska Glassmeier, Tom Goren, Daniel~P Grosvenor, et~al.
\newblock Opportunistic experiments to constrain aerosol effective radiative forcing.
\newblock \emph{Atmospheric chemistry and physics}, 22\penalty0 (1):\penalty0 641--674, 2022.

\bibitem[Cohen et~al.(2017)Cohen, Brauer, Burnett, Anderson, Frostad, Estep, Balakrishnan, Brunekreef, Dandona, Dandona, et~al.]{cohen2017estimates}
Aaron~J Cohen, Michael Brauer, Richard Burnett, H~Ross Anderson, Joseph Frostad, Kara Estep, Kalpana Balakrishnan, Bert Brunekreef, Lalit Dandona, Rakhi Dandona, et~al.
\newblock Estimates and 25-year trends of the global burden of disease attributable to ambient air pollution: an analysis of data from the global burden of diseases study 2015.
\newblock \emph{The lancet}, 389\penalty0 (10082):\penalty0 1907--1918, 2017.

\bibitem[Conover(1966)]{10.1175/1520-0469(1966)}
John~H. Conover.
\newblock {Anomalous Cloud Lines}.
\newblock \emph{Journal of the Atmospheric Sciences}, 23\penalty0 (6):\penalty0 778--785, 1966.
\newblock ISSN 0022-4928.
\newblock \doi{10.1175/1520-0469(1966)023<0778:acl>2.0.co;2}.

\bibitem[Deng et~al.(2009)Deng, Dong, Socher, Li, Li, and Fei-Fei]{deng2009imagenet}
Jia Deng, Wei Dong, Richard Socher, Li-Jia Li, Kai Li, and Li~Fei-Fei.
\newblock Imagenet: A large-scale hierarchical image database.
\newblock In \emph{2009 IEEE conference on computer vision and pattern recognition}, pages 248--255. Ieee, 2009.

\bibitem[Forster et~al.(2021)Forster, Storelvmo, Armour, Collins, Dufresne, Frame, Lunt, Mauritsen, Palmer, Watanabe, Wild, and Zhang]{IPCC}
P.~Forster, T.~Storelvmo, K.~Armour, W.~Collins, J.~L. Dufresne, D.~Frame, D.~J. Lunt, T.~Mauritsen, M.~D. Palmer, M.~Watanabe, M.~Wild, and H.~Zhang.
\newblock {The Earth’s Energy Budget, Climate Feedbacks, and Climate Sensitivity}.
\newblock Technical report, Cambridge University Press, 2021.

\bibitem[Goldman et~al.(2019)Goldman, Herzig, Eisenschtat, Goldberger, and Hassner]{goldman2019precise}
Eran Goldman, Roei Herzig, Aviv Eisenschtat, Jacob Goldberger, and Tal Hassner.
\newblock Precise detection in densely packed scenes.
\newblock In \emph{Proceedings of the IEEE/CVF conference on computer vision and pattern recognition}, pages 5227--5236, 2019.

\bibitem[Goren and Rosenfeld(2012)]{10.1029/2012JD017981}
Tom Goren and Daniel Rosenfeld.
\newblock {Satellite observations of ship emission induced transitions from broken to closed cell marine stratocumulus over large areas}.
\newblock \emph{Journal of Geophysical Research: Atmospheres}, 117\penalty0 (D17):\penalty0 n/a--n/a, 9 2012.
\newblock ISSN 0148-0227.
\newblock \doi{10.1029/2012JD017981}.

\bibitem[Gryspeerdt et~al.(2019)Gryspeerdt, Smith, O'Keeffe, Christensen, and Goldsworth]{10.1029/2019GL084700}
Edward Gryspeerdt, Tristan W.~P. Smith, Eoin O'Keeffe, Matthew~W. Christensen, and Fraser~W. Goldsworth.
\newblock {The Impact of Ship Emission Controls Recorded by Cloud Properties}.
\newblock \emph{Geophysical Research Letters}, 46\penalty0 (21):\penalty0 12547--12555, 11 2019.
\newblock ISSN 0094-8276.
\newblock \doi{10.1029/2019GL084700}.

\bibitem[He et~al.(2016)He, Zhang, Ren, and Sun]{he2016deep}
Kaiming He, Xiangyu Zhang, Shaoqing Ren, and Jian Sun.
\newblock Deep residual learning for image recognition.
\newblock In \emph{Proceedings of the IEEE conference on computer vision and pattern recognition}, pages 770--778, 2016.

\bibitem[He et~al.(2017)He, Gkioxari, Doll{\'a}r, and Girshick]{he2017mask}
Kaiming He, Georgia Gkioxari, Piotr Doll{\'a}r, and Ross Girshick.
\newblock Mask r-cnn.
\newblock In \emph{Proceedings of the IEEE international conference on computer vision}, pages 2961--2969, 2017.

\bibitem[Heguerte et~al.(2023)Heguerte, Bugeau, and Lannelongue]{heguerte2023estimate}
Lucia~Bouza Heguerte, Aur{\'e}lie Bugeau, and Lo{\"\i}c Lannelongue.
\newblock How to estimate carbon footprint when training deep learning models? a guide and review.
\newblock \emph{arXiv preprint arXiv:2306.08323}, 2023.

\bibitem[Huang et~al.(2017)Huang, Liu, Van Der~Maaten, and Weinberger]{huang2017densely}
Gao Huang, Zhuang Liu, Laurens Van Der~Maaten, and Kilian~Q Weinberger.
\newblock Densely connected convolutional networks.
\newblock In \emph{Proceedings of the IEEE conference on computer vision and pattern recognition}, pages 4700--4708, 2017.

\bibitem[Keipour et~al.(2022)Keipour, Mousaei, Bandari, Schaal, and Scherer]{keipour2022detection}
Azarakhsh Keipour, Mohammadreza Mousaei, Maryam Bandari, Stefan Schaal, and Sebastian Scherer.
\newblock Detection and physical interaction with deformable linear objects.
\newblock \emph{arXiv preprint arXiv:2205.08041}, 2022.

\bibitem[Lin et~al.(2014)Lin, Maire, Belongie, Hays, Perona, Ramanan, Doll{\'a}r, and Zitnick]{lin2014microsoft}
Tsung-Yi Lin, Michael Maire, Serge Belongie, James Hays, Pietro Perona, Deva Ramanan, Piotr Doll{\'a}r, and C~Lawrence Zitnick.
\newblock Microsoft coco: Common objects in context.
\newblock In \emph{Computer Vision--ECCV 2014: 13th European Conference, Zurich, Switzerland, September 6-12, 2014, Proceedings, Part V 13}, pages 740--755. Springer, 2014.

\bibitem[Lin et~al.(2017)Lin, Doll{\'a}r, Girshick, He, Hariharan, and Belongie]{lin2017feature}
Tsung-Yi Lin, Piotr Doll{\'a}r, Ross Girshick, Kaiming He, Bharath Hariharan, and Serge Belongie.
\newblock Feature pyramid networks for object detection.
\newblock In \emph{Proceedings of the IEEE conference on computer vision and pattern recognition}, pages 2117--2125, 2017.

\bibitem[Ng et~al.(2023)Ng, McCloskey, Cui, Meijer, Brand, Sarna, Goyal, Arsdale, and Geraedts]{ng2023opencontrails}
Joe Yue-Hei Ng, Kevin McCloskey, Jian Cui, Vincent~R. Meijer, Erica Brand, Aaron Sarna, Nita Goyal, Christopher~Van Arsdale, and Scott Geraedts.
\newblock Opencontrails: Benchmarking contrail detection on goes-16 abi, 2023.

\bibitem[Ronneberger et~al.(2015)Ronneberger, Fischer, and Brox]{ronneberger2015u}
Olaf Ronneberger, Philipp Fischer, and Thomas Brox.
\newblock U-net: Convolutional networks for biomedical image segmentation.
\newblock In \emph{Medical Image Computing and Computer-Assisted Intervention--MICCAI 2015: 18th International Conference, Munich, Germany, October 5-9, 2015, Proceedings, Part III 18}, pages 234--241. Springer, 2015.

\bibitem[Schwartz et~al.(2020)Schwartz, Dodge, Smith, and Etzioni]{schwartz2020green}
Roy Schwartz, Jesse Dodge, Noah~A Smith, and Oren Etzioni.
\newblock Green ai.
\newblock \emph{Communications of the ACM}, 63\penalty0 (12):\penalty0 54--63, 2020.

\bibitem[Shindell and Smith(2019)]{shindell2019climate}
Drew Shindell and Christopher~J Smith.
\newblock Climate and air-quality benefits of a realistic phase-out of fossil fuels.
\newblock \emph{Nature}, 573\penalty0 (7774):\penalty0 408--411, 2019.

\bibitem[Tan and Le(2019)]{tan2019efficientnet}
Mingxing Tan and Quoc Le.
\newblock Efficientnet: Rethinking model scaling for convolutional neural networks.
\newblock In \emph{International conference on machine learning}, pages 6105--6114. PMLR, 2019.

\bibitem[Twomey et~al.(1968)Twomey, Howell, and Wojciechowski]{twomey1968comments}
S~Twomey, HB~Howell, and TA~Wojciechowski.
\newblock Comments on “anomalous cloud lines”.
\newblock \emph{Journal of the Atmospheric Sciences}, 25\penalty0 (2):\penalty0 333--334, 1968.

\bibitem[Wada(2018)]{Wada_Labelme_Image_Polygonal}
Kentaro Wada.
\newblock {Labelme: Image Polygonal Annotation with Python}, 2018.
\newblock URL \url{https://github.com/wkentaro/labelme}.

\bibitem[Wang et~al.(2022)Wang, Zhang, Cui, Ren, Yang, Xie, Hua, Bao, and Xu]{wang2022active}
Chi Wang, Yunke Zhang, Miaomiao Cui, Peiran Ren, Yin Yang, Xuansong Xie, Xian-Sheng Hua, Hujun Bao, and Weiwei Xu.
\newblock Active boundary loss for semantic segmentation.
\newblock In \emph{Proceedings of the AAAI Conference on Artificial Intelligence}, volume~36, pages 2397--2405, 2022.

\bibitem[Wang et~al.(2020)Wang, Zhang, Kong, Li, and Shen]{wang2020solov2}
Xinlong Wang, Rufeng Zhang, Tao Kong, Lei Li, and Chunhua Shen.
\newblock Solov2: Dynamic and fast instance segmentation.
\newblock \emph{Advances in Neural information processing systems}, 33:\penalty0 17721--17732, 2020.

\bibitem[Watson-Parris and Smith(2022)]{10.1038/s41558-022-01516-0}
Duncan Watson-Parris and Christopher~J. Smith.
\newblock {Large uncertainty in future warming due to aerosol forcing}.
\newblock \emph{Nature Climate Change}, pages 1--3, 2022.
\newblock ISSN 1758-678X.
\newblock \doi{10.1038/s41558-022-01516-0}.

\bibitem[Watson-Parris et~al.(2022)Watson-Parris, Christensen, Laurenson, Clewley, Gryspeerdt, and Stier]{watson2022shipping}
Duncan Watson-Parris, Matthew~W Christensen, Angus Laurenson, Daniel Clewley, Edward Gryspeerdt, and Philip Stier.
\newblock Shipping regulations lead to large reduction in cloud perturbations.
\newblock \emph{Proceedings of the National Academy of Sciences}, 119\penalty0 (41):\penalty0 e2206885119, 2022.

\bibitem[Xie et~al.(2017)Xie, Girshick, Doll{\'a}r, Tu, and He]{xie2017aggregated}
Saining Xie, Ross Girshick, Piotr Doll{\'a}r, Zhuowen Tu, and Kaiming He.
\newblock Aggregated residual transformations for deep neural networks.
\newblock In \emph{Proceedings of the IEEE conference on computer vision and pattern recognition}, pages 1492--1500, 2017.

\bibitem[Yuan et~al.(2022)Yuan, Song, Wood, Wang, Oreopoulos, Platnick, Hippel, Meyer, Light, and Wilcox]{10.1126/sciadv.abn7988}
Tianle Yuan, Hua Song, Robert Wood, Chenxi Wang, Lazaros Oreopoulos, Steven~E. Platnick, Sophia~von Hippel, Kerry Meyer, Siobhan Light, and Eric Wilcox.
\newblock {Global reduction in ship-tracks from sulfur regulations for shipping fuel}.
\newblock \emph{Science Advances}, 8\penalty0 (29):\penalty0 eabn7988, 2022.
\newblock \doi{10.1126/sciadv.abn7988}.

\bibitem[Zhu et~al.(2019)Zhu, Hu, Lin, and Dai]{zhu2019deformable}
Xizhou Zhu, Han Hu, Stephen Lin, and Jifeng Dai.
\newblock Deformable convnets v2: More deformable, better results.
\newblock In \emph{Proceedings of the IEEE/CVF conference on computer vision and pattern recognition}, pages 9308--9316, 2019.

\bibitem[Ångström and Angstrom(1929)]{10.2307/519399.Ångström}
Anders Ångström and Anders Angstrom.
\newblock {On the Atmospheric Transmission of Sun Radiation and on Dust in the Air}.
\newblock \emph{Geografiska Annaler}, 11:\penalty0 156, 1929.
\newblock ISSN 1651-3215.
\newblock \doi{10.2307/519399}.

\end{thebibliography}
